\documentclass[sigconf, nonacm=true]{acmart} 
\AtBeginDocument{%
  \providecommand\BibTeX{{%
    \normalfont B\kern-0.5em{\scshape i\kern-0.25em b}\kern-0.8em\TeX}}}

\setcopyright{acmcopyright}
\copyrightyear{2021}
\acmYear{2021}
\acmDOI{10.1145/1122445.1122456}

\acmConference[SIGKDD '21]{SIGKDD '21: ACM SIG on Knowledge Discovery and Data Mining}{August 14--18, 2021}{Virtual Event, Singapore}
\acmBooktitle{Proceedings of the 27th ACM SIGKDD Conference on Knowledge Discovery and Data Mining (KDD ’21), August 14--18, 2021, Virtual Event, Singapore}

\acmSubmissionID{1948}

\usepackage{microtype}
\usepackage{todonotes}
\usepackage{graphicx}
\usepackage{subfigure}
\usepackage{booktabs} 
\usepackage{hyperref}
\usepackage{comment}

\usepackage{amsmath}
\usepackage{adjustbox}
\usepackage{bbm}
\usepackage{ctable}
\usepackage{multirow}
\usepackage[ruled,vlined]{algorithm2e}

\DeclareMathOperator*{\argmax}{argmax}
\DeclareMathOperator*{\argmin}{argmin}

\begin{document}

\title{Ensemble Squared: A Meta AutoML System}

\author{Jason Yoo}
\affiliation{%
  \institution{University of British Columbia}
  \city{Vancouver}
  \country{Canada}
}
\email{jasony97@cs.ubc.ca}

\author{Tony Joseph}
\affiliation{%
  \institution{University of British Columbia}
  \city{Vancouver}
  \country{Canada}
}
\email{tonyjos@cs.ubc.ca}

\author{Dylan Yung}
\affiliation{%
  \institution{Georgia Institute of Technology}
  \city{Atlanta}
  \country{United States of America}
}
\email{dyung6@gatech.edu}

\author{S. Ali Nasseri}
\affiliation{%
  \institution{University of British Columbia}
  \city{Vancouver}
  \country{Canada}
}
\email{ali.nasseri@ubc.ca}

\author{Frank Wood}
\affiliation{%
  \institution{University of British Columbia}
  \city{Vancouver}
  \country{Canada}
}
\email{fwood@cs.ubc.ca}

\renewcommand{\shortauthors}{Yoo et al.}

\begin{abstract}
There are currently many barriers that prevent non-experts from exploiting machine learning solutions ranging from the lack of intuition on statistical learning techniques to the trickiness of hyperparameter tuning. Such barriers have led to an explosion of interest in automated machine learning (AutoML), whereby an off-the-shelf system can take care of many of the steps for end-users without the need for expertise in machine learning. This paper presents Ensemble Squared (Ensemble$^2$), an AutoML system that ensembles the results of state-of-the-art open-source AutoML systems. Ensemble$^2$ exploits the diversity of existing AutoML systems by leveraging the differences in their model search space and heuristics. Empirically, we show that diversity of each AutoML system is sufficient to justify ensembling at the AutoML system level. In demonstrating this, we also establish new state-of-the-art AutoML results on the OpenML tabular classification benchmark.
\end{abstract}



\maketitle

\section{Introduction}

Advances in computer hardware and the ever-expanding abundance of data are enabling the construction of machine learning models that are yielding progressively more value in a growing set of application domains. Unfortunately, it is well known that there is no single best machine learning model that can handle all problems \cite{Wolpert1996}. As such, for every new application area or problem, a laborious, largely manual process must be followed that includes data cleaning, feature engineering, testing and model design. This is what data scientists do, often in collaboration with domain experts. With the demand for data science talent on the job market exceeding supply  \cite{manyika2011big, pompa2017data, analytics2016age}, the societal ``value add'' from machine learning is bottlenecked by the availability of data scientists and the complexity of applied machine learning methods.

One solution to this problem is to automate much of the machine learning model development and pipeline selection process using automated machine learning (AutoML) techniques. AutoML systems allow data scientists to focus directly on value creation, i.e.~developing methods for solving the underlying problem, while other tasks such as data cleaning, model selection, model fitting, and hyperparameter tuning are automated. The full promise of AutoML is that, eventually, it will enable non-data scientists to exploit performant predictive models and extract insights from data.

A variety of AutoML systems exist today \citep{autogluon, feurer2015efficient, heffetz2019deepline, H2OAutoML20} 
and have begun to coalesce around reporting results on the OpenML classification benchmark \cite{gijsbers2019open}. Our review of the open-source AutoML systems and their results led us to ask the following question: Ensembling individual models has been shown to be an extremely powerful idea \citep{hansen1990ensemble}; could it be that there is sufficient diversity in AutoML systems so that ensembling AutoML systems would lead to quantitative gains? 

This is the question that this paper addresses. Ensemble$^2$, the system we built, ensembles machine learning pipelines searched for in parallel by a set of diverse AutoML systems (our base systems), and achieves a new state of the art baseline on the OpenML classification benchmark. To our knowledge, no other existing work ensembles results among different AutoML systems, exploiting the different heuristics between them in terms of pipeline search, model selection, and hyperparameter optimization strategies to achieve better performance.

\begin{figure*}[t]
   \begin{center}
   \includegraphics[width=\textwidth]{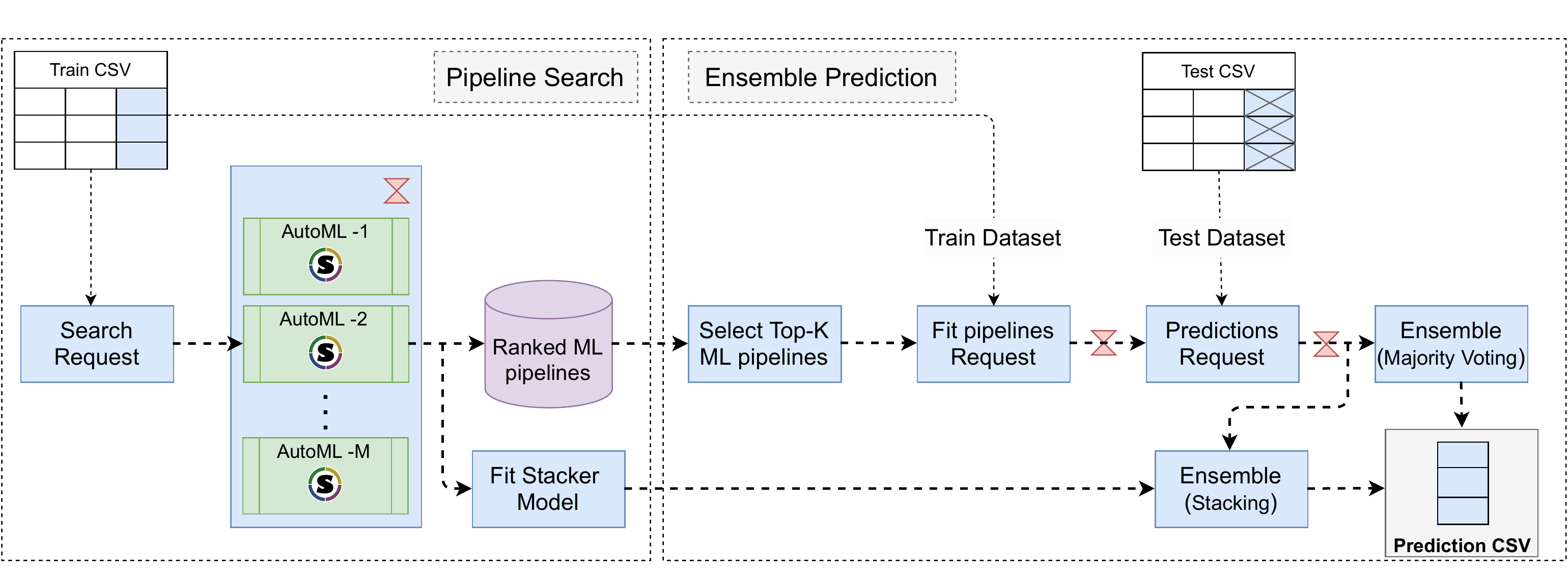}
   \end{center}
   \caption[Ensemble$^2$ workflow]{Overview of Ensemble$^2$ workflow. {\normalfont Ensemble$^2$ is made up of two underlying subsystems. The training dataset (in CSV format), along with a user-specified target column, is sent to the \textit{Pipeline Search Subsystem}. This spins up $M$ different AutoML systems (our base systems) as Singularity containers and performs the pipeline search procedure in parallel for a set time duration. Once the time limit has been hit and all discovered pipelines ($P$) have been collected, it ranks the pipelines based on their validation scores. The ranked pipelines, alongside the  training and test datasets are then passed on to the second subsystem, the \textit{Ensemble Prediction Subsystem}. Here, the top $K$ pipelines (where $K$ is an Ensemble$^2$ hyperparameter and $K \leq P$), are selected, optionally refit, and made to generate predictions on the test dataset. The predictions are then passed onto the ensembling module, which generates the final Ensemble$^2$ predictions either using majority voting or a stacking model. The stacking model can choose to ensemble the best pipeline from each AutoML system as well.}}
   \label{fig:main}
\end{figure*}

We are aware that numerous AutoML solutions already {\em internally} ensemble machine learning pipelines discovered during their internal search phase \citep{feurer2015efficient, feurer2020auto, autogluon, chen2018autostacker, zaidi2020nes}. They do so for the same reason that we explore ensembling AutoML systems {\em externally}: ensembling can reduce both bias and variance of a machine learning models' predictions. 

Ensemble$^2$ has an additional benefit of being more robust than its base systems. Existing AutoML systems are brittle, and can fail to return solutions on some datasets due to: out-of-memory errors, missing values, missing metafeatures, incorrect identification of task, and incorrectly data typed columns \cite{zoller2019benchmark}. This is strongly linked to the differences in pre-processing steps, and the machine learning techniques implemented in them that can have different bugs, and varying degrees of functional flexibility. Ensemble$^2$, by its construction, is more robust than all of its underlying AutoML systems and only fails when all base systems fail.

Figure \ref{fig:main} presents an overview of Ensemble$^2$. Our system consists of two subsystems, one that performs pipeline search using the base systems in parallel and another that ensembles results.  Tables~\ref{tab:v1result1hr} and ~\ref{tab:v2result1hr} show Ensemble$^2$'s performance on the OpenML classification benchmark datasets relative to the AutoML systems it ensembles. We find that Ensemble$^2$ achieves the highest average rank overall and that it outperforms notable current open-source AutoML systems such as: AutoGluon\cite{autogluon}, Auto-Sklearn \cite{feurer2015efficient}, Auto-Sklearn 2.0 \cite{feurer2020auto}, CMU AutoML \footnote{\url{https://github.com/autonlab/cmu-ta2}}, and H2O AutoML \cite{H2OAutoML20}.

In addition to our results, we have built a public-facing web interface that enables anyone to upload their own CSV and use Ensemble$^2$. Ensemble$^2$ is also easily scalable by the virtue of containerization using Singularity \cite{singularity} and integration with Slurm~\cite{Yoo2003}.

\section{Background} 
\label{section:background}

Ensemble$^2$ relies on ensembling the best pipelines from five base AutoML systems: H2O AutoML \cite{H2OAutoML20}, Auto-Sklearn \cite{feurer2015efficient}, Auto-Sklearn 2.0 \cite{feurer2020auto}, AutoGluon \cite{autogluon}, and CMU AutoML which was developed as part of the DARPA D3M program. These systems were selected for their differing heuristics and search spaces. Table~\ref{comparison} contains a brief overview of these base systems and the different approaches they take to the automate the pipeline search and hyperparameter optimization problem.

\begin{table*}[t]
\centering
\def\arraystretch{1.30}
\begin{adjustbox}{width=\textwidth}
\begin{tabular}{|p{2.5cm}||p{2.5cm}|p{4.5cm}|p{4.5cm}|}
 \specialrule{.2em}{.1em}{.1em}
 \hline
 AutoML System & Primitive Library & \begin{tabular}[c]{@{}c@{}}Model Discovery and \\ Hyperparameter Tuning \end{tabular} & Internal Ensembling \\
 \hline
 \hline
  AutoGluon & Gluon Library & Fixed Defaults & Multi-Layer Stacking and Bagging \\
 \hline
  Auto-SkLearn & Scikit-Learn  & BayesOpt + Meta-Learning  & Forward Search\\
 \hline
  Auto-SkLearn 2.0 & Scikit-Learn & Portfolio Learning  & Forward Search \\
 \hline
 CMU AutoML & D3M Primitives & Templates + Grid Search & -\\
 \hline
  H2O AutoML & H2O Library & Grid and Random Search & Super Learner\\
 \hline
\end{tabular}
\end{adjustbox}
\caption{Base AutoML Systems used in Ensemble$^2$. This table highlights some of the differences between these AutoML systems and the diversity of methods that Ensemble$^2$ benefits from.}
\label{comparison}
\end{table*}

\subsection{Problem Formulation}

While there are several definitions of the AutoML problem, our work follows the definition in \citet{zoller2019benchmark}. Let a machine learning pipeline $P: \mathcal{X} \rightarrow \mathcal{Y}$ be a sequential combination of algorithms that transforms a feature vector $x \in \mathcal{X}$ to a target value $y \in \mathcal{Y}$. For example, $y$ is an one-hot vector of class labels for a classification problem and a real number for a one-dimensional regression problem. Let $\mathcal{A} = \{A^{(1)},A^{(2)},...,A^{(n)}\}$ be a fixed set of data-cleaning, feature extraction, and estimator algorithms where each algorithm $A^{(i)}$ is configured by a set of hyperparameters $\lambda^{(i)}$ from the domain $\Lambda^{(i)}$. Then, $P$'s structure can be described as a Directed Acyclic Graph (DAG) where each node is an algorithm $A^{(i)}$ and each edge represents the data flow between algorithms.

The objective of an AutoML system is to find the configuration of algorithms and hyperparameters that minimizes the loss on an unseen test dataset. Ensemble$^2$ estimates $P^*$ by ensembling (Section~\ref{sec:ensembleLearning}) the returned pipelines by its base AutoML systems, where $P^*$ is defined as: 
\begin{equation}
    \label{eq:obj}
    P^* = \mathrm{argmin}_{P \in \mathcal{P}}\; \mathcal{L}(\mathcal{D}_{train}, \mathcal{D}_{test}, P)
\end{equation}
where $P$ is a valid pipeline from the space of all valid DAG-structured pipelines $\mathcal{P}$, $\mathcal{L}$ is the task loss, and $\mathcal{D}_{train}$ and $\mathcal{D}_{test}$ are the training and test datasets respectively.  

\subsection{Ensemble Learning} \label{sec:ensembleLearning}

Ensembling methods are commonly used to boost performance by reducing model bias and variance compared to the base learners \cite{rahman2014ensemble}. Popular ensembling methods often utilize one of voting, bagging \cite{Breiman1996}, boosting \cite{Freund1996}, or stacking \cite{Wolpert1992} techniques. Voting involves having the base learners vote on the correct class with equal weight and taking the class with the most votes as the final prediction. Bagging involves independently training the base learners using a randomly drawn subset of the training set and having the base learners vote with equal weight. Boosting involves incrementally constructing an ensemble by training base learners to better classify training data points that the previous base learners misclassified. Lastly, stacking involves training a classifier that generates predictions by looking at the predictions of its base learners. An instance of stacking paradigm is the Super Learner algorithm \cite{SuperLearner}, which trains its stacker on out-of-fold predictions from the base learners. 

In this paper, we explored majority voting and stacking strategies (Section~\ref{sec:ensembleMethods}). In \textit{majority voting}, it selects the top-$K$ pipelines based on the lowest cross-validation loss, where $K$ is an integer value set by the user. For this work and the web interface backend, we have set $K$ to be three.
In the second strategy, \textit{stacking}, the top-1 pipeline returned from every base AutoML systems are used to learn a stacking model.

\vspace{-1em}

\subsection{Base AutoML Systems}
As previously mentioned, five base AutoML systems were used in Ensemble$^2$. These systems were selected as the base AutoML systems because of their strong individual performance shown in \cite{zoller2019benchmark} as well as our own empirical results. We explored other state-of-the-art AutoML systems such as GAMA \cite{gijsbers2019open}, TPOT \citep{olson2016tpot} (which uses genetic programming), DeepLine \cite{heffetz2019deepline}, AlphaD3M \cite{drori2019automatic} (which both use reinforcement learning), and Auto-Weka 2.0 \cite{JMLR:v18:16-261} (which uses Bayesian optimization), but settled on the chosen five based on performance and ease of integration. 

The different search strategies are elaborated below, so we only detail their different search spaces in this paragraph. Auto-Sklearn and Auto-Sklearn 2.0 restrict their ML algorithms (ex. data-cleaning, feature pre-processing, and estimator) to the scikit-learn library \cite{sckitlearn}. CMU AutoML restricts its ML algorithms the the D3M library. AutoGluon uses a mix of scikit-learn library, neural networks, and custom-made tree-based ML algorithm. Lastly, H2O AutoML uses its custom pre-built models consisting of random forest, gradient boosting machines, linear and deep learning models.\\

\noindent \textbf{AutoGluon.} $\;$ AutoGluon \cite{autogluon} ensembles pre-built classification models that have been shown to do empirically well in classification tasks. AutoGluon employs a unique stacking technique called Multi-Layer Stack Ensembling alongside K-fold ensemble bagging. Multi-Layer Stack Ensembling involves stacking models in multiple layers and training in a layer-wise manner which would guarantee high-quality predictions within a given time frame. To prevent overfitting, Autogluon relies on K-fold ensemble bagging at all layers.\\

\noindent \textbf{Auto-Sklearn.} $\;$ Auto-Sklearn \cite{feurer2015efficient} uses Bayesian optimization to search through the model and hyperparameter space of selected scikit-learn \cite{sckitlearn} modules. Meta-learning is used to warm start the search procedure. When the model search is over, Auto-Sklearn constructs an ensemble from discovered pipelines by employing forward search \cite{caruana2004ensemble}.
\\

\noindent \textbf{Auto-Sklearn 2.0} $\;$ Auto-Sklearn 2.0 \cite{feurer2020auto} improves upon Auto-Sklearn by employing portfolio learning. It first runs Bayesian optimization on all meta-train datasets and constructs a portfolio of pipelines to run on all future tasks. In addition, it learns what model selection strategy to use on a new dataset based on simple meta-features involving some combination of k-fold validation, validation holdout set, and successive halving \cite{jamieson2015sh}. It also uses various other improvements such as intermittent result storage to be able to return results quicker for larger datasets in a short amount of time.
\\

\noindent \textbf{CMU AutoML.} $\;$ CMU AutoML takes a template approach to the pipeline synthesis problem. More specifically, it uses multiple hand-crafted pre-processing steps and cycles through all possible pre-processing and estimator combinations to quickly assess the optimality of many pipelines in a short amount of time. This follows from the authors' finding that foregoing hyperparameter search and instead using that time to evaluate as many models as possible was able to achieve superior performance when dealing with short search times. However, some key pipeline component hyperparameters are grid-searched.
\\

\noindent \textbf{H2O AutoML.} $\;$ H2O AutoML \cite{H2OAutoML20} ensembles variations of the followings models: random forest, gradient boosting machines, linear models, deep learning and stacked iterations of the aforementioned models. The AutoML system does a 5-fold cross validation scheme for training and evaluating each model and takes the model with the highest cross validation score (logloss for classification, mean squared error for regression) across all 5-folds and refits the top ranked model on the entire train data set. This refit model is then returned and used for prediction. All models have a predefined range for each hyperparameter defined as "most important", which are hard coded into the AutoML system. Random search is then performed on the finite set of hyperparameters to find the optimal models. H2O AutoML creates two stacked models (Super Learner models) to evaluate against the other models, one that ensembles all models and one that ensembles the best model of each of the aforementioned types.

\section{Methodology}

In this section, we will discuss the two ensembling approaches we have experimented with for tackling tabular classification problems: \textit{Majority Voting} and \textit{Super Learner}~\cite{SuperLearner}.

\subsection{Ensemble Methods} \label{sec:ensembleMethods}

\noindent \textbf{Majority Voting.} $\;$ Let us define the feature vector space as $\mathcal{X}$, and the $|C|$-dimensional one-hot target vector space as $\mathcal{Y}$. We denote $P: \mathcal{X} \rightarrow \mathcal{Y}$ to be a machine learning pipeline with fully defined algorithm and hyperparameter values. Let $\mathbf{P} = \{P_1, P_2, ..., P_J\}$ be a set of $J$ machine learning pipelines from the base AutoML systems that achieves the lowest cross-validation loss when the search procedure is over. Lastly, we denote the indicator function as $\mathbb{I}$. 

A majority voting ensemble model $M: \mathcal{X} \rightarrow \mathcal{Y}$ assigns the label $\hat{y}$ to a data point $x$ as follows:
\begin{equation}
    \hat{y} = \argmax_{y} \left(\sum_{i=1}^N \mathbb{I}(\mathbf{P}_i(x) = y)\right)
\end{equation}
where $\mathbf{P}_i$ are trained on $\mathcal{D}_{train}$ with the base classifier's loss. The advantage of the majority voting approach is the fact that no additional training is required aside from training the pipelines in $\mathbf{P}$.
\\



\noindent \textbf{Super Learner.} $\;$ The Super Learner algorithm is a stacking algorithm that fits a stacking model $M$ on the out-of-fold predictions from the base learners $\mathbf{P}$. At training time, the algorithm first partitions the training dataset $\mathcal{D}_{train}$ into $K$ mutually exclusive folds and fits each of its $J$ base learners on all $K$ folds. Then, each base learner produces an out-of-fold prediction on the unseen parts of the training data. Lastly, a stacking model of choice $M$ learns a weighting over the predictions from all base learners.

A formal definition of $M$'s training objective is as follows:
\begin{equation}
    \theta = \argmin_{\theta} \mathcal{L}(M, \mathcal{D}_{train}, \mathbf{P},\theta)
\end{equation}
where $\theta$ is $M$'s parameters, $\mathbf{P}$ is a set of $J \times K$ base learners fit on different folds of $\mathcal{D}_{train}$, and $\mathcal{L}$ is a minimizer over any given loss function of the parameter space of $\theta$.


Ensemble$^2$ uses a softmax regression model for $M$. However, we note that $M$ can be any classifier or even an artifact from another AutoML system. Given all the base learner predictions on a single datapoint $x$, the softmax regression model $M: \mathcal{J} \rightarrow \mathcal{Y}$ assigns the label $\hat{y}$ to $x$ as follows:

\begin{equation}
    \label{eqn:softreghyp}
    \hat{y} = \argmax_{y} \left( \frac{exp(\theta_y^Tx)}{\sum_{c=1}^{C}exp(\theta^T_{c}x)} \right)
\end{equation}
where $c\in\{1,..,C\}$ and $\theta_c \in R^J$ are the learned coefficients during the training phase.





\section{Architecture}



Ensemble$^2$ employs a minimalist API that interfaces with D3M gRPC API \footnote{\url{https://gitlab.com/datadrivendiscovery/ta3ta2-api}}, Auto-Sklearn API \footnote{\url{https://automl.github.io/auto-sklearn/master/index.html}}, AutoGluon API, and H2O API. Its design is highly extendable. All one needs to do to add an AutoML system is creating a Singularity container that can run on \textit{search} and \textit{predict} mode. It also allows for a client-server setup, where the AutoML systems run on server mode and Ensemble$^2$ can schedule various commands to the AutoML systems from the client side all within a single Singularity container.

Ensemble$^2$'s containerization approach enables us to run the AutoML systems locally or remotely on clusters using a workload manager like Slurm. Scaling Ensemble$^2$ for production use is as simple as specifying which Slurm partition(s) Ensemble$^2$ should run on. A single Ensemble$^2$ run can trivially be deployed across many nodes because of its parallel design with respect to its base systems. Containerization using Singularity also helps us overcome the specific OS-dependencies that some base AutoML systems have.

Such a minimalist client-server API interface allowed us to build a web-based user-interface, which end-users can use without expertise in machine learning or programming. The user can directly upload their training dataset, select the target column, and specify the search time. Once search request is submitted, the \textit{pipeline search} stage starts and the user will be notified after the search is completed. The user will then be able to upload their test dataset and send in the prediction request, which will start \textit{ensemble prediction} stage. Once this stage is done, the user will be able to download the final prediction dataset.

\subsection{Pipeline Search Stage}
Given a training CSV, Ensemble$^2$ spins up a Singularity container for each of its base AutoML system in parallel. This is done by sending search requests to each base AutoML system to perform the pipeline search procedure for a given duration of time with a specific seed. Discovered ML pipelines along with their cross validation scores and out-of-fold predictions are saved to disk. If Super Learner algorithm is selected, Ensemble$^2$ fits the softmax regression stacker on out-of-fold predictions generated by the base AutoML systems.

\subsection{Ensembling Stage}
Given a test CSV, Ensemble$^2$ first collects all discovered pipelines and ranks them based on their search time validation scores. The top-$K$ pipelines (where $K$ is an Ensemble$^2$ hyperparameter) are then selected and their respective AutoML systems are spin-up in parallel to generate predictions on the test CSV. After the predictions are generated, Ensemble$^2$ generates its final test dataset predictions by top-$K$ majority voting or by passing the individual predictions to the stacking model $M$.

\section{Empirical Evaluation}

We experimented with two versions of Ensemble$^2$. Version-1 (V1) ensembles AutoGluon, Auto-Sklearn, Auto-Sklearn 2.0, while Version-2 (V2) ensembles AutoGluon, Auto-Sklearn, Auto-Sklearn 2.0, CMU AutoML, and H2O AutoML. This enables us to investigate if adding more AutoML systems improves the overall perfomance and robustness of Ensemble$^2$. V1 and V2 both perform majority voting and Super Learner ensembles. The sections below describe the evaluation datasets, different experiment schemes we have performed, and their respective results.

\subsection{Datasets}

All experiments were performed on the OpenML classification benchmark datasets \cite{gijsbers2019open}. The benchmark was curated such that this set of datasets vary in the number of data points and features by orders of magnitudes. Each dataset also varies in the number of categorical features, numerical features, and missing values. The benchmark datasets also do not contain classification problems that are too easy to solve (e.g. most artificially generated datasets) and is gradually updated overtime to prevent AutoML tools from overfitting to the member datasets. At the time of this experiment, there were 41 datasets in the benchmark. The evaluation metric is accuracy for all datasets in the benchmark.

\subsection{Evaluating Ensemble$^2$ V1}

This experiment compares the accuracy of the predictions generated by the base systems against the Ensemble$^2$'s prediction, which are generated by ensembling the aforementioned predictions. We call this the wall-clock experiment, since all systems are given the same amount of time of one hour for pipeline search.
\\

\noindent \textbf{Setup.} $\;$ For each benchmark dataset, all AutoML systems were run in parallel for the time limit. Every system had access to 4 CPUs and 8GB of RAM, and were performed across machines with Intel E5-2683 v4 Broadwell, 2.1GHz processor and Intel Xeon Gold 5120 Skylake, 2.2GHz processor. Ensemble$^2$'s majority voting ensembled the top three pipelines with the best validation scores from the base AutoML systems.

We used the default settings for AutoGluon, Auto-Sklearn, and Auto-Sklearn 2.0 with the following exceptions. AutoGluon used its best performance mode ($auto\_stack = True$) since we wanted performance over resource efficiency for benchmarking purposes. Auto-Sklearn used cross-validation as its resampling strategy with number of folds set to 3 and had its memory limit set to 8 GB. Auto-Sklearn 2.0's configuration was identical to Auto-Sklearn except for its re-sampling strategy which it determined based on the dataset meta-features. Both Auto-Sklearn and Auto-Sklearn 2.0 had 25\% of their search times allocated to refitting their best discovered models on the entire training dataset as they recommend refitting on the whole dataset at the end when using cross-validation re-sampling strategy. Allocating less than 25\% of search time for refitting led to failures on certain datasets.

For Super Learner, the best models from each AutoML system were used, mainly because the majority of the AutoML systems only produced out-of-fold predictions for their best models. When a model could not produce out-of-fold predictions they were excluded from the Super Learner. An example of this is Auto-Sklearn 2.0, which uses a different evaluation method depending on the dataset and only generates out-of-fold predictions when cross-validation is chosen. Lastly, we applied L2-penalty when training our softmax regression stacking model.

To account for scenarios where AutoML systems didn't terminate on time, we gave each Singularity container 30 minutes of grace period to smoothly exit the search and the predict phase. Despite this precaution, we found that rarely AutoML systems did not terminate even after the grace period but saved the model weights on disk that can be used for prediction. In this case, we loaded the saved models at prediction phase and counted those runs as successes. If there were multiple saved models in such a manner, a random model was selected and scored. 

We also observed quite a bit of out-of-memory errors during our initial runs, so each Singularity container was given an extra 4 GB of RAM to prevent it from crashing. Lastly, in cases when AutoML runs still failed, we re-ran them one more time with identical configurations to ensure that the failures weren't sporadic. Note, such a measure was taken for the accurate reporting of scores generated by the base AutoML systems. In a production scenario, a non-expert using our system would still receive a model, if even one of the base AutoML systems successfully completed the search. A failure would occur only if all the base AutoML systems failed.
\\

\begin{table*}[p]
\centering
\def\arraystretch{1.25}
\begin{adjustbox}{width=0.8\textwidth, height=0.45\textheight}
\begin{tabular}{|c|>{\hspace{1pc}}c>{\hspace{1pc}}c>{\hspace{1pc}}c|c|c|}
\specialrule{.2em}{.1em}{.1em}
\hline
  \textbf{\begin{tabular}[c]{@{}c@{}}OpenML \\ Dataset ID \\  \end{tabular}} &
  \textbf{\begin{tabular}[c]{@{}c@{}}AutoGluon \\             \end{tabular}} &
  \textbf{\begin{tabular}[c]{@{}c@{}}Auto-Sklearn \\         \end{tabular}} &
  \textbf{\begin{tabular}[c]{@{}c@{}}Auto-Sklearn 2.0 \\         \end{tabular}} &
  \textbf{\begin{tabular}[c]{@{}c@{}}Ensemble$^2$\\ Voting    \end{tabular}} &
  \textbf{\begin{tabular}[c]{@{}c@{}}Ensemble$^2$\\ Stacking  \end{tabular}} \\ 
\hline
2     & 0.989±0.001  & \textbf{0.993}±0.001  & 0.989±0.002  & 0.991±0.002     & 0.989±0.001       \\
3     & 0.993±0.002  & 0.996±0.001  & \textbf{0.997}±0.000  & 0.994±0.001     & 0.993±0.002       \\
5     & 0.684±0.012  & \textbf{0.740}±0.013  & 0.736±0.010  & 0.707±0.019     & 0.673±0.012       \\
12    & 0.975±0.002  & \textbf{0.984}±0.003  & 0.977±0.001  & 0.978±0.004     & 0.975±0.002       \\
31    & \textbf{0.779}±0.003  & 0.772±0.003  & 0.778±0.004  & 0.778±0.002     & \textbf{0.779}±0.003       \\
54    & 0.829±0.016  & 0.845±0.010  & 0.798±0.016  & \textbf{0.846}±0.011     & 0.825±0.014       \\
1067  & \textbf{0.859}±0.003  & 0.856±0.004  & 0.854±0.003  & \textbf{0.859}±0.004     & \textbf{0.859}±0.003       \\
1111  & \textbf{0.983}±0.000  & \textbf{0.983}±0.000  & \textbf{0.983}±0.000  & \textbf{0.983}±0.000     & \textbf{0.983}±0.000       \\
1169  & 0.632±0.014  & 0.669±0.000  & \textbf{0.670}±0.001  & 0.667±0.004     & 0.576±0.041       \\
1461  & \textbf{0.908}±0.001  & 0.906±0.000  & 0.905±0.001  & \textbf{0.908}±0.001     & \textbf{0.908}±0.001       \\
1464  & 0.751±0.012  & \textbf{0.761}±0.008  & 0.742±0.009  & 0.751±0.011     & 0.751±0.012       \\
1468  & 0.918±0.008  & \textbf{0.950}±0.006  & 0.947±0.013  & 0.936±0.017     & 0.918±0.008       \\
1486  & \textbf{0.973}±0.001  & \textbf{0.973}±0.001  & 0.971±0.002  & \textbf{0.973}±0.001     & \textbf{0.973}±0.001       \\
1489  & 0.902±0.002  & \textbf{0.904}±0.002  & 0.902±0.002  & 0.902±0.002     & 0.902±0.002       \\
1590  & \textbf{0.876}±0.001  & 0.875±0.001  & \textbf{0.876}±0.001  & \textbf{0.876}±0.001     & \textbf{0.876}±0.001       \\
1596  & 0.889±0.006  & \textbf{0.969}±0.000  & 0.968±0.001  & 0.951±0.031     & 0.889±0.006       \\
4135  & \textbf{0.950}±0.000  & 0.949±0.000  & 0.949±0.001  & 0.949±0.001     & \textbf{0.950}±0.000       \\
23512 & 0.727±0.001  & \textbf{0.731}±0.000  & 0.729±0.000  & 0.727±0.001     & 0.727±0.001       \\
23517 & 0.509±0.002  & \textbf{0.520}±0.000  & \textbf{0.520}±0.001  & 0.502±0.004     & 0.509±0.002       \\
40668 & 0.839±0.007  &\textbf{0.849}±0.000  & 0.658±0.000  & 0.843±0.005     & 0.839±0.007       \\
40685 & \textbf{1.000}±0.000  & \textbf{1.000}±0.000  & \textbf{1.000}±0.000  & \textbf{1.000}±0.000     & \textbf{1.000}±0.000       \\
40975 & 0.979±0.002  & 0.981±0.002  & \textbf{0.984}±0.001  & 0.979±0.003     & 0.979±0.002       \\
40981 & 0.869±0.006  & 0.869±0.010  & \textbf{0.870}±0.005  & 0.868±0.008     & 0.869±0.006       \\
40984 & \textbf{0.941}±0.001  & 0.935±0.001  & 0.940±0.001  & 0.940±0.001     & \textbf{0.941}±0.001       \\
40996 & \textbf{0.900}±0.002  & 0.717±0.309  & 0.731±0.315  & \textbf{0.900}±0.002     & \textbf{0.900}±0.002       \\
41027 & \textbf{0.966}±0.002  & 0.910±0.020  & 0.862±0.001  & \textbf{0.966}±0.002     & \textbf{0.966}±0.002       \\
41138 & \textbf{0.994}±0.000  & 0.993±0.000  & 0.990±0.000  & \textbf{0.994}±0.000     & \textbf{0.994}±0.000       \\
41142 & 0.736±0.003  & \textbf{0.750}±0.004  & 0.737±0.002  & 0.736±0.003     & 0.736±0.003       \\
41143 & 0.804±0.002  & \textbf{0.820}±0.006  & 0.808±0.000  & 0.804±0.002     & 0.804±0.002       \\
41146 & \textbf{0.949}±0.002  & 0.948±0.004  & 0.944±0.001  & \textbf{0.949}±0.002     & \textbf{0.949}±0.002       \\
41147 & 0.679±0.005  & 0.615±0.062  & \textbf{0.689}±0.000  & 0.684±0.006     & 0.679±0.005       \\
41150 & \textbf{0.948}±0.001  & 0.945±0.000  & \textbf{0.948}±0.000  & \textbf{0.948}±0.001     & \textbf{0.948}±0.001       \\
41159 & \textbf{0.821}±0.006  & 0.643±0.051  & 0.643±0.051  & \textbf{0.821}±0.005     & \textbf{0.821}±0.006       \\
41161 & \textbf{0.998}±0.000  & 0.997±0.000  & 0.997±0.000  & \textbf{0.998}±0.001     & \textbf{0.998}±0.000       \\
41163 & \textbf{0.989}±0.001  & 0.986±0.002  & 0.978±0.001  & \textbf{0.989}±0.001     & \textbf{0.989}±0.001       \\
41164 & \textbf{0.723}±0.002  & 0.714±0.004  & 0.714±0.004  & \textbf{0.723}±0.002     & \textbf{0.723}±0.002       \\
41165 & 0.478±0.015  & \textbf{0.480}±0.000  & \textbf{0.480}±0.000  & 0.477±0.015     & 0.478±0.015       \\
41166 & \textbf{0.716}±0.001  & 0.127±0.000  & 0.700±0.000  & 0.710±0.001     & \textbf{0.716}±0.001       \\
41167 & \textbf{0.913}±0.003  & 0.005±0.000  & 0.005±0.000  & \textbf{0.913}±0.003     & 0.612±0.428       \\
41168 & 0.719±0.001  & 0.706±0.000  & \textbf{0.720}±0.000  & 0.718±0.001     & 0.719±0.001       \\
41169 & 0.259±0.080  & 0.007±0.000  & 0.007±0.000  & \textbf{0.297}±0.058     & 0.144±0.146       \\ \hline
{\begin{tabular}[c]{@{}c@{}}Average Accuracy \\ \end{tabular}} & $0.838$ & $0.790$ & $0.797$ & $\textbf{0.842}$ & $0.826$\\ \hline
{\begin{tabular}[c]{@{}c@{}}Average Rank \\ \end{tabular}} & $2.878$ & $3.024$ & $3.268$ & $\textbf{2.817}$ & $3.012$\\ \hline
{\begin{tabular}[c]{@{}c@{}}\# First Place \\ \end{tabular}} & $\textbf{20}$ & $16$ & $12$ & $18$ & $19$\\ \hline
\end{tabular}%
\end{adjustbox}
\caption{Comparison between AutoGluon, Auto-Sklearn, Auto-Sklearn 2.0 and Ensemble$^2$'s ensemble of their results. All systems were run for one hour and the results were averaged across five seeds. Failures were omitted during the mean calculations. Ensemble$^2$'s majority voting setup ensembles top three pipelines irrespective of which AutoML systems they come from while the Super Learning setup ensembles the best pipelines returned by each system. The highest accuracy achieved by a method on a dataset is shown in boldface.}
\label{tab:v1result1hr}
\end{table*}

\noindent \textbf{Result.} $\;$ All base system scores were measured by the performances of the single best pipeline. We note that most of these systems already employ ensembling internally, and the returned pipelines are often already ensembled. Ensemble$^2$'s majority voting score was computed from the predictions generated by ensembling the top three pipelines with the highest validation accuracy from any of the base systems. Ensemble$^2$'s Super Learner score was computed from the predictions generated by ensembling the best pipelines returned by each successful AutoML system. Failures were removed for average calculation.

Table \ref{tab:v1result1hr} summarizes the results of this experiment. The average rank of each AutoML system relative to one another is written in the bottom row of the table, with ties allowed. Overall, Ensemble$^2$'s majority voting approach achieved the highest average accuracy and rank of $0.842$ and $2.817$ across five seeds, which is slightly higher than the second highest average accuracy and rank of $0.838$ and $2.878$ achieved by AutoGluon. The Super Learner approach did not perform as well as the majority voting approach, with the average accuracy and rank of $0.826$ and $3.012$. We suspect that the Super Learner didn't perform as well because it only received predictions from three base learners. In future experiments, we plan to experiment with increasing the number of base learners to assess their impact on performance.

To ensure that the difference in performance distribution over the datasets in the benchmark between pairs of AutoML systems was statistically significant, we ran Wilcoxon signed-rank test with $\alpha = 0.05$ by setting the results of each AutoML systems on unique dataset and seed combinations as datapoints. Since Ensemble$^2$'s majority voting scheme achieved first place, we computed the test statistic between majority voting Ensemble$^2$ with every other AutoML system. The resulting p-values for rejecting the null hypothesis that the pair produced the same results all indicated statistical significance. Specifically, the p-values for AutoGluon, Auto-Sklearn, and Auto-Sklearn 2.0 were $0.0008$, $0.0473$, and $10^{-5}$.

We noticed that AutoGluon achieves first place on more datasets than Ensemble$^2$'s voting scheme in Table \ref{tab:v1result1hr} ($20/41$ vs $18/41$ datasets) even though Ensemble$^2$ had both higher average accuracy and rank compared to AutoGluon. When we computed the number of times an AutoML system achieves first place only between AutoGluon and Ensemble$^2$, we discovered that Ensemble$^2$ had first place on $33$ of the $41$ datasets and Autogluon had first place on $30$ out of the $41$ datasets. This entails that Ensemble$^2$ does have more wins over AutoGluon across benchmark datasets, but the datasets on which Ensemble$^2$ performs better can have other AutoML systems that perform better than both Ensemble$^2$ and AutoGluon.
\\

\begin{table}[t]
\begin{center}
\tabcolsep=5pt
\small
\begin{tabular}{ |c||c|c|c| }
 \specialrule{.2em}{.1em}{.1em}
 \hline
 \textbf{System} & \textbf{Avg. Accuracy} & \textbf{Avg. Rank} & \textbf{\# 1st Place} \\
 \hline
 AutoGluon        & 0.841 & 2.878 & 15 \\ 
 Auto-Sklearn     & 0.790 & \textbf{2.536} & \textbf{18} \\ 
 Auto-Sklearn 2.0 & 0.797 & 3.134 & 15 \\ 
 Ensemble$^2$ Voting & \textbf{0.842} & 3.146 & 10 \\ 
 Ensemble$^2$ Stacking & 0.826 & 3.304 & 10 \\ 
 \hline
\end{tabular}
\end{center}
\caption{Summary statistics for 1 hour Ensemble$^2$ V1 runs vs three hour base system runs. Voting and Stacking denote Ensemble$^2$'s majority voting and Super Learner implementations respectively.}
\label{tab:v1equalcompute}
\end{table}

\noindent \textbf{Equal-Compute Comparison.} $\;$ While the wall-clock experiment can show that Ensemble$^2$ produces superior results in the same time-frame, Ensemble$^2$ has access to more compute than its competitors under this setup. To investigate how Ensemble$^2$ fairs when it has access to the same amount of compute power as its base systems, we compared the performance of 1 hour run Ensemble$^2$ against three hour run base AutoML systems. The comparison results are listed on Table \ref{tab:v1equalcompute}. The reported three hour performances were averaged across five seeds.

While Ensemble$^2$'s voting scheme retained the highest average accuracy, AutoGluon's average accuracy increased to the point where Ensemble$^2$'s performance gain is insignificant. In addition, while Auto-Sklearn's average accuracy decreased, it had more overall wins over other AutoML systems. 

\subsection{Evaluating Ensemble$^2$ V2}

\noindent \textbf{Setup.} $\;$ The setup for Ensemble$^2$ V2 is identical to the setup for Ensemble$^2$ V1 except for some key differences. Firstly, the AutoML systems were run across four machines with Intel Core i7-5820K, 3.3 GHz processors, and 48 GB DDR4 RAM. Secondly, both CMU AutoML and H2O AutoML used cross-validation as their resampling strategy, with number of folds set to 3 and 5 respectively. H2O AutoML's allowed number of threads was set to 4 and maximum memory size was set to 8 GB.
\\

\noindent \textbf{Result.} $\;$ Table \ref{tab:v2result1hr} summarizes the results of this experiment. Again, Ensemble$^2$'s majority voting approach achieved the highest average accuracy and rank of $0.844$ and $2.963$ across five seeds, which is slightly higher than the second highest average rank of $3.085$ achieved by AutoGluon.

\begin{table}[h]
\begin{center}
\tabcolsep=5pt
\small
\begin{tabular}{|c||c|c|c|}
 \specialrule{.2em}{.1em}{.1em}
 \hline
 \textbf{System} & \textbf{Avg. Accuracy} & \textbf{Avg. Rank} & \textbf{\# 1st Place} \\
 \hline
 AutoGluon           & \textbf{0.848} & \textbf{2.183} & \textbf{19} \\ 
 Auto-Sklearn        & 0.783          & 5.183          & 5 \\ 
 Auto-Sklearn 2.0    & 0.804          & 4.695          & 4 \\ 
 CMU AutoML          & 0.806          & 4.720          & 7 \\ 
 H2O AutoML          & 0.793          & 4.976          & 1 \\ 
 Ensemble$^2$ Voting           & 0.844          & 2.902          & 9 \\ 
 Ensemble$^2$ Stacking         & 0.840          & 3.341          & 5 \\ 
 \hline
\end{tabular}
\end{center}
\caption{Summary statistics for 1 hour Ensemble$^2$ V2 runs vs five hour base system runs. Voting and Stacking denote Ensemble$^2$'s majority voting and Super Learner implementations respectively.}
\label{tab:v2equalcompute}
\end{table}

We ran the same Wilcoxon signed-rank test with $\alpha = 0.05$ to ensure the performance differences between Ensemble$^2$ majority voting and base AutoML systems were significant. The resulting p-values for rejecting the null hypothesis that the pair produced the same results all indicated statistical significance. Specifically, the p-values for AutoGluon, Auto-Sklearn, and Auto-Sklearn 2.0 were $0.0048$, $0.0117$, and $0.0001$. The p-values for CMU AutoML and H2O AutoML were less than $10^{-5}$.

\begin{table*}[p]
\centering
\def\arraystretch{1.28}
\begin{adjustbox}{width=\textwidth}
\begin{tabular}{|c||>{\hspace{1pc}}c>{\hspace{1pc}}c>{\hspace{1pc}}c>{\hspace{1pc}}c>{\hspace{1pc}}c|c|c|}
\specialrule{.2em}{.1em}{.1em}
\hline
  \textbf{\begin{tabular}[c]{@{}c@{}}OpenML \\ Dataset ID \\ \end{tabular}} &
  \textbf{\begin{tabular}[c]{@{}c@{}}AutoGluon \\ \end{tabular}} &
  \textbf{\begin{tabular}[c]{@{}c@{}}Auto-Sklearn \\ \end{tabular}} &
  \textbf{\begin{tabular}[c]{@{}c@{}}Auto-Sklearn 2.0 \\ \end{tabular}} &
  \textbf{\begin{tabular}[c]{@{}c@{}}CMU AutoML \\ \end{tabular}} &
  \textbf{\begin{tabular}[c]{@{}c@{}}H2O AutoML \\ \end{tabular}} &
  \textbf{\begin{tabular}[c]{@{}c@{}}Ensemble$^2$\\ Voting \end{tabular}} &
  \textbf{\begin{tabular}[c]{@{}c@{}}Ensemble$^2$\\ Stacking \end{tabular}} \\ \hline
2     & 0.989±0.000  & 0.992±0.002  & 0.992±0.000  & 0.992±0.000 & \textbf{0.993}±0.001 & \textbf{0.993}±0.001     & 0.990±0.002       \\
3     & 0.994±0.001  & \textbf{0.997}±0.001  & \textbf{0.997}±0.000  & 0.985±0.022 & 0.992±0.003 & 0.993±0.002     & 0.993±0.002       \\
5     & 0.687±0.006  & 0.734±0.006  & \textbf{0.745}±0.009  & 0.696±0.064 & 0.727±0.022 & 0.725±0.020     & 0.685±0.023       \\
12    & 0.973±0.002  & \textbf{0.983}±0.002  & 0.972±0.005  & 0.975±0.004 & 0.967±0.004 & 0.977±0.003     & 0.973±0.003       \\
31    & 0.778±0.004  & \textbf{0.781}±0.005  & 0.777±0.005  & 0.757±0.020 & 0.752±0.006 & 0.776±0.004     & 0.777±0.005       \\
54    & 0.833±0.017  & \textbf{0.841}±0.003  & 0.798±0.012  & 0.783±0.007 & 0.820±0.007 & 0.835±0.011     & 0.829±0.014       \\
1067  & 0.854±0.003  & \textbf{0.857}±0.004  & 0.852±0.002  & 0.853±0.005 & 0.805±0.016 & 0.854±0.003     & 0.856±0.004       \\
1111  & \textbf{0.983}±0.000  & \textbf{0.983}±0.000  & \textbf{0.983}±0.000  & \textbf{0.983}±0.000 & 0.972±0.004 & \textbf{0.983}±0.000     & \textbf{0.983}±0.000       \\
1169  & 0.658±0.005  & 0.653±0.030  & \textbf{0.670}±0.000  & 0.634±0.040 & 0.633±0.005 & 0.658±0.005     & 0.658±0.005       \\
1461  & \textbf{0.908}±0.001  & 0.906±0.000  & 0.906±0.001  & 0.903±0.002 & 0.905±0.002 & \textbf{0.908}±0.001     & \textbf{0.908}±0.001       \\
1464  & 0.751±0.012  & 0.761±0.007  & 0.745±0.005  & \textbf{0.770}±0.012 & 0.737±0.031 & 0.764±0.012     & 0.747±0.009       \\
1468  & 0.917±0.005  & 0.946±0.004  & \textbf{0.952}±0.003  & 0.819±0.064 & 0.941±0.002 & 0.949±0.005     & 0.930±0.008       \\
1486  & \textbf{0.973}±0.000  & 0.971±0.000  & 0.954±0.006  & 0.958±0.014 & 0.971±0.001 & \textbf{0.973}±0.000     & \textbf{0.973}±0.000       \\
1489  & 0.902±0.002  & \textbf{0.906}±0.002  & 0.905±0.001  & 0.902±0.003 & 0.892±0.003 & 0.902±0.002     & 0.902±0.002       \\
1590  & \textbf{0.876}±0.001  & \textbf{0.876}±0.000  & 0.875±0.000  & 0.834±0.026 & \textbf{0.876}±0.001 & \textbf{0.876}±0.000     & \textbf{0.876}±0.001       \\
1596  & 0.899±0.036  & \textbf{0.969}±0.000  & 0.968±0.000  & 0.674±0.159 & 0.783±0.227 & 0.968±0.002     & 0.899±0.036       \\
4135  & 0.949±0.001  & 0.949±0.000  & \textbf{0.950}±0.000  & 0.946±0.002 & 0.947±0.001 & 0.949±0.001     & 0.949±0.001       \\
23512 & 0.726±0.001  & \textbf{0.730}±0.002  & 0.728±0.003  & 0.683±0.035 & 0.722±0.002 & 0.726±0.001     & 0.725±0.001       \\
23517 & 0.510±0.002  & \textbf{0.520}±0.000  & 0.519±0.001  & 0.518±0.001 & 0.507±0.001 & 0.507±0.005     & 0.510±0.002       \\
40668 & 0.828±0.007  & 0.848±0.000  & 0.725±0.081  & 0.797±0.032 & \textbf{0.853}±0.004 & 0.851±0.005     & \textbf{0.853}±0.004       \\
40685 & \textbf{1.000}±0.000  & 0.958±0.084  & 0.958±0.084  & 0.920±0.094 & \textbf{1.000}±0.000 & \textbf{1.000}±0.000     & \textbf{1.000}±0.000       \\
40975 & 0.979±0.002  & 0.982±0.001  & \textbf{0.984}±0.001  & 0.960±0.020 & 0.978±0.004 & 0.980±0.002     & 0.980±0.004       \\
40981 & \textbf{0.867}±0.005  & 0.866±0.002  & \textbf{0.867}±0.011  & 0.865±0.005 & 0.859±0.007 & 0.862±0.012     & 0.862±0.006       \\
40984 & \textbf{0.940}±0.002  & 0.935±0.002  & 0.938±0.002  & 0.908±0.045 & 0.932±0.003 & \textbf{0.940}±0.003     & \textbf{0.940}±0.002       \\
40996 & \textbf{0.897}±0.001  & 0.829±0.093  & 0.887±0.001  & 0.761±0.161 & 0.861±0.010 & \textbf{0.897}±0.001     & 0.895±0.001       \\
41027 & \textbf{0.965}±0.002  & 0.914±0.019  & 0.862±0.001  & 0.907±0.081 & 0.861±0.002 & \textbf{0.965}±0.002     & \textbf{0.965}±0.002       \\
41138 & \textbf{0.994}±0.000  & 0.993±0.000  & 0.990±0.000  & 0.993±0.000 & 0.993±0.001 & \textbf{0.994}±0.000     & \textbf{0.994}±0.000       \\
41142 & 0.736±0.002  & \textbf{0.749}±0.006  & 0.737±0.002  & 0.700±0.038 & 0.721±0.003 & 0.735±0.002     & 0.736±0.002       \\
41143 & 0.802±0.003  & 0.809±0.005  & \textbf{0.810}±0.002  & 0.794±0.010 & 0.791±0.008 & 0.803±0.002     & 0.789±0.010       \\
41146 & 0.948±0.001  & 0.944±0.002  & 0.945±0.001  & 0.948±0.009 & 0.941±0.003 & \textbf{0.949}±0.001     & 0.945±0.004       \\
41147 & \textbf{0.689}±0.001  & 0.565±0.000  & 0.687±0.005  & 0.631±0.000 & 0.662±0.001 & 0.688±0.002     & \textbf{0.689}±0.001       \\
41150 & 0.947±0.001  & 0.944±0.002  & \textbf{0.948}±0.000  & 0.946±0.001 & 0.944±0.001 & 0.947±0.001     & 0.945±0.001       \\
41159 & \textbf{0.819}±0.001  & 0.749±0.054  & 0.813±0.012  & 0.813±0.012 & 0.783±0.014 & \textbf{0.819}±0.001     & \textbf{0.819}±0.001       \\
41161 & \textbf{0.997}±0.000  & \textbf{0.997}±0.000  & \textbf{0.997±}0.000  & \textbf{0.997}±0.000 & 0.947±0.060 & \textbf{0.997}±0.000     & \textbf{0.997}±0.000       \\
41163 & \textbf{0.989}±0.002  & 0.983±0.000  & 0.975±0.002  & 0.975±0.002 & 0.968±0.007 & 0.988±0.002     & \textbf{0.989}±0.001       \\
41164 & \textbf{0.723}±0.003  & 0.714±0.006  & 0.711±0.004  & 0.701±0.011 & 0.692±0.006 & \textbf{0.723}±0.003     & 0.722±0.003       \\
41165 & 0.480±0.016  & 0.481±0.002  & \textbf{0.482}±0.012  & \textbf{0.482}±0.012 & 0.380±0.002 & 0.479±0.015     & 0.474±0.015       \\
41166 & \textbf{0.715}±0.001  & 0.608±0.118  & 0.693±0.006  & 0.620±0.038 & 0.668±0.003 & 0.710±0.000     & 0.714±0.001       \\
41167 & \textbf{0.871}±0.047  & 0.570±0.344  & 0.703±0.318  & 0.240±0.000 & 0.300±0.003 & 0.856±0.066     & 0.870±0.048       \\
41168 & \textbf{0.718}±0.001  & 0.706±0.000  & \textbf{0.718}±0.002  & 0.697±0.014 & 0.712±0.002 & \textbf{0.718}±0.001     & 0.715±0.003       \\
41169 & \textbf{0.396}±0.001  & 0.353±0.027  & 0.371±0.002  & 0.345±0.022 & 0.312±0.005 & \textbf{0.396}±0.001     & 0.391±0.001       \\ \hline
{\begin{tabular}[c]{@{}c@{}}Average Accuracy\\ \end{tabular}} & $0.840$ & $0.826$ & $0.831$ & $0.807$ & $0.797$ & $\textbf{0.844}$ & $0.840$\\ \hline
{\begin{tabular}[c]{@{}c@{}}Average Rank\\ \end{tabular}} & $3.085$ & $3.585$ & $3.634$ & $5.707$ & $5.427$ & $\textbf{2.963}$ & $3.598$\\ \hline
{\begin{tabular}[c]{@{}c@{}}\# First Place \\ \end{tabular}} & $\textbf{19}$ & $13$ & $13$ & $4$ & $4$ & $16$ & $13$\\ \hline
\end{tabular}%
\end{adjustbox}
\caption{Comparison between AutoGluon, Auto-Sklearn, Auto-Sklearn 2.0, CMU AutoML, H2O AutoML, and Ensemble$^2$'s ensemble of their results. All systems were run for one hour and the results were averaged across five seeds. Failures were omitted during the mean calculations. Ensemble$^2$'s majority voting setup ensembles top three pipelines irrespective of which AutoML systems they come from while the Super Learning setup ensembles the best pipelines returned by each system. The highest accuracy achieved by a method on a dataset is shown in boldface.}
\label{tab:v2result1hr}
\end{table*}

We noticed again that AutoGluon achieves first place on more datasets than Ensemble$^2$'s voting scheme in Table \ref{tab:v1result1hr} ($19/41$ vs $16/41$ datasets) even though Ensemble$^2$ had both higher average accuracy and rank compared to AutoGluon. When we computed the number of times an AutoML system achieves first place only between AutoGluon and Ensemble$^2$, we discovered that Ensemble$^2$ had first place on $31$ of the $41$ datasets and AutoGluon had first place on $30$ out of the $41$ datasets.
\\

\noindent \textbf{Equal-Compute Comparison.} $\;$ To investigate how Ensemble$^2$ V2 fairs when it has access to the same amount of compute power as its base systems, we compared the performance of 1 hour run Ensemble$^2$ against five hour run base AutoML systems. The comparison results are listed on Table \ref{tab:v2equalcompute}. The reported five hour performances were computed from a single seed due to time constraints.

We found that overall, AutoGluon performed best under these conditions. Curiously, the average accuracy for all other AutoML systems slightly dropped on five hour runs compared to 1 hour runs. This is likely due to the fact that running AutoML search for long time risks overfitting, and AutoGluon takes extensive measures to counteract that by using tools like repeated $k$-fold bagging.

\subsection{AutoML System Performance Correlation}

\begin{figure}[ht]
  \center
  \includegraphics[scale=0.38]{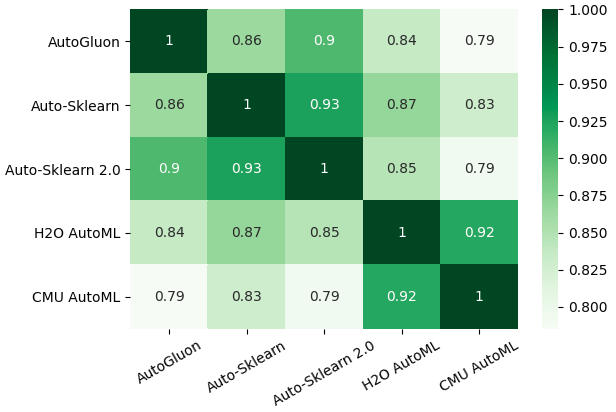}
  \caption{Correlation in test set performance between Ensemble$^2$'s base AutoML systems.}
  \label{fig:correlation}
\end{figure}

We investigate whether Ensemble$^2$'s base AutoML systems perform similarly across benchmark tasks by looking at the correlation between the accuracy of the base systems. This correlation measures whether the base systems generally perform better on different kinds of datasets. The less correlated the base systems are, the more suited they are for different kinds of datasets. Ensemble$^2$ generally benefits from having a suite of lowly correlated base systems since it can rely on at least one of the base systems to do well on a wide range of problems. Figure \ref{fig:correlation} shows that some diversity exists in the performance of the best pipelines generated by Ensemble$^2$'s base systems. For example, the correlation between AutoGluon and CMU AutoML is $\sim$0.79 and the correlation between H2O AutoML and Auto-Sklearn 2.0 is $\sim$0.85. Hence, we argue that Ensemble$^2$ is overall more well-rounded than its base systems.



\section{Conclusion}
In this paper, we have established that ensembling AutoML systems generally lead to quantitative gains in accuracy. We have observed that Ensemble$^2$ outperformed all SOTA AutoML systems under wall-clock comparisons. On equal-compute comparisons, we saw that Ensemble$^2$'s performance was competitive to SOTA AutoML systems. This, and our explicit examination of the correlation between the accuracy of various AutoML systems across a variety of problems, suggests that there is, currently, exploitable diversity between AutoML systems.

\sloppy In addition to those findings, we have built a public-facing Ensemble$^2$ web interface designed for simplicity and ease of scaling. This web system greatly increases the accessibility of machine learning for the general public, as a non-data scientist user can simply upload a training CSV, specify which column to classify, and finally uploads a test CSV to receive predictions.

Our experiments had a one-hour search time because that is the most common timeout that the majority of AutoML papers use for their experiments. To assess how Ensemble$^2$'s performance gain fluctuates, for both wall-clock and equal-compute setups, it would be valuable to try running experiments with very short and very long search times. In addition to experimenting with different search times, it would be informative to ensemble many more combinations of AutoML systems while taking into account their search space and heuristics to observe what combinations yield the best empirical results. We have observed that some AutoML systems had lower average accuracy than others, so a choice of better performing AutoML systems and taking measures to further prevent overfitting could also improve the performance of the ensembling system compared to a single system like AutoGluon on equal compute setup.

\begin{acks}
We acknowledge the support of the Natural Sciences and Engineering Research Council of Canada (NSERC), the Canada CIFAR AI Chairs Program, and the Intel Parallel Computing Centers program. This material is based upon work supported by the United States Air Force Research Laboratory (AFRL) under the Defense Advanced Research Projects Agency (DARPA) Data Driven Discovery Models (D3M) program (Contract No. FA8750-19-2-0222) and Learning with Less Labels (LwLL) program (Contract No.FA8750-19-C-0515). Additional support was provided by UBC's Composites Research Network (CRN), Data Science Institute (DSI) and Support for Teams to Advance Interdisciplinary Research (STAIR) Grants. This research was enabled in part by technical support and computational resources provided by WestGrid (https://www.westgrid.ca/) and Compute Canada (www.computecanada.ca).
\end{acks}

\bibliographystyle{ACM-Reference-Format}
\bibliography{main-ref}




\end{document}